\documentclass[runningheads]{llncs}
\usepackage[T1]{fontenc}
\usepackage{graphicx,verbatim}
\usepackage{amsmath,amssymb,amsfonts}
\usepackage{booktabs}
\usepackage{multirow}
\usepackage{array}
\usepackage{makecell}
\usepackage{subfigure}
\usepackage{bm}
\usepackage{marvosym}
\usepackage[lined,boxed,commentsnumbered,ruled]{algorithm2e}
\usepackage[colorlinks,
            linkcolor=blue,
            anchorcolor=blue,
            citecolor=blue]{hyperref}
\begin{document}
\title{High-Order Progressive Trajectory Matching for Medical Image Dataset Distillation}

\author{Le Dong\inst{1} \and
Jinghao Bian\inst{1}\thanks{Work done during an internship at MedAI Technology (Wuxi) Co. Ltd.} \and
Jingyang Hou\inst{2} \and
Jingliang Hu\inst{2} \and
Yilei Shi\inst{2} \and
\\Weisheng Dong\inst{1} \and
Xiao Xiang Zhu\inst{3} \and
Lichao Mou\inst{2}\textsuperscript{(\Letter)}}

% index{Dong, Le}
% index{Bian, Jinghao}
% index{Hou, Jingyang}
% index{Hu, Jingliang}
% index{Shi, Yilei}
% index{Dong, Weisheng}
% index{Zhu, Xiao Xiang}
% index{Mou, Lichao}

\authorrunning{L. Dong et al.}
% \institute{Anonymous Organization\\
% \email{**@******.***}}

\institute{Xidian University, Xi’an, China \and MedAI Technology (Wuxi) Co. Ltd., Wuxi, China\\\email{lichao.mou@medimagingai.com} \and Technical University of Munich, Munich, Germany}
%

% \author{Anonymized Authors}  %% Added for anonymized MICCAI 2025 submission
% \authorrunning{Anonymized Author et al.}
% \institute{Anonymized Affiliations \\
%     \email{email@anonymized.com}}

\maketitle              % typeset the header of the contribution
\begin{abstract}
Medical image analysis faces significant challenges in data sharing due to privacy regulations and complex institutional protocols. Dataset distillation offers a solution to address these challenges by synthesizing compact datasets that capture essential information from real, large medical datasets. Trajectory matching has emerged as a promising methodology for dataset distillation; however, existing methods primarily focus on terminal states, overlooking crucial information in intermediate optimization states. We address this limitation by proposing a shape-wise potential that captures the geometric structure of parameter trajectories, and an easy-to-complex matching strategy that progressively addresses parameters based on their complexity. Experiments on medical image classification tasks demonstrate that our method improves distillation performance while preserving privacy and maintaining model accuracy comparable to training on the original datasets. Our code is available at \url{https://github.com/Bian-jh/HoP-TM}.

\keywords{patient privacy \and dataset distillation \and trajectory matching \and image classification.}
% Authors must provide keywords and are not allowed to remove this Keyword section.

\end{abstract}
\section{Introduction}
\label{sec:intro}
Deep learning has revolutionized medical image analysis, achieving unprecedented performance across diverse tasks. However, developing deep networks in the medical domain faces a critical challenge: the scarcity of large-scale, diverse training datasets. This limitation arises from multiple domain-specific constraints, including stringent privacy regulations~\cite{mcgraw2009privacy,shen2019understanding} and complex data sharing protocols across healthcare institutions~\cite{weitzman2010sharing,jin2019review}. While approaches like federated learning~\cite{rieke2020future,xu2021federated,tian2023communication} address privacy concerns, they demand substantial computational resources and intricate coordination among participating institutions.
\par
Dataset distillation offers a promising solution to these challenges by synthesizing a small set of training samples that encapsulate the essential information of a large-scale dataset~\cite{wang2018dataset}. This approach creates a compact, synthetic dataset that, when used for training, yields models with performance comparable to those trained on the original, full dataset. In the medical imaging context, dataset distillation not only addresses privacy concerns by avoiding direct data sharing but also significantly reduces storage requirements and computational costs associated with model training~\cite{li2020soft,li2022compressed,li2022dataset}.
\par
Recent years have witnessed the emergence of various algorithmic frameworks for dataset distillation, including gradient matching methods that align gradients produced by synthetic and real data~\cite{kim2022dataset,zhao2021dataset}, kernel inducing points using neural tangent kernel ridge regression~\cite{nguyen2020dataset,nguyen2021dataset}, and distribution matching approaches that minimize the distance between synthetic and real data distributions~\cite{wang2022cafe,zhao2023dataset,zhao2023improved}. Among these, trajectory matching~\cite{cazenavette2022dataset,du2023minimizing,guo2023towards,liu2025dataset} has demonstrated superior performance. This approach leverages pre-recorded expert trajectories from models trained on real data and optimizes synthetic examples to reproduce similar parameter trajectories.
\par
Despite its promise, existing trajectory matching methods primarily focus on matching terminal states, essentially comparing only the destinations of parameter trajectories~\cite{cazenavette2022dataset,du2023minimizing,guo2023towards,liu2025dataset}. This simplified approach overlooks crucial information embedded in intermediate optimization states. To address this limitation, we propose matching trajectories with a high-order potential. Specifically, we introduce a shape-wise potential that captures the geometric structure of parameter trajectories, enabling more comprehensive matching.
\par
Moreover, we observe that model parameters exhibit varying degrees of matching complexity during optimization. Building on this insight, we develop an easy-to-complex matching strategy that progressively addresses parameter matching based on complexity. This approach initially matches simpler parameters before advancing to more challenging ones, resulting in robust and effective trajectory matching.
\par
Our contributions are three-fold:
\begin{itemize}
\item We propose a shape-wise potential that enables high-order trajectory matching, capturing rich geometric relationships between parameter trajectories.
\item We introduce an easy-to-complex matching strategy that enhances trajectory matching by considering parameter-specific characteristics.
\item A thorough investigation of dataset distillation in medical imaging, with extensive experiments demonstrating state-of-the-art performance.
\end{itemize}

\section{Method}
\subsection{Preliminary}
Trajectory matching methods~\cite{cazenavette2022dataset,du2023minimizing,guo2023towards,liu2025dataset} first use multiple teacher networks trained for $T$ epochs on a real dataset to generate a set of expert trajectories $\{\bm{\tau}\}$, where each trajectory $\bm{\tau}$ represents a time sequence of parameters ${\{\bm{\theta}_t\}}_{0}^{T}$. 
\par
At each distillation iteration, we randomly sample start parameters $\bm{\theta}_t$ and target parameters $\bm{\theta}_{t+M}$ ($t+M\leq T$) from an expert trajectory. A student network, initialized with $\bm{\theta}_t$, is then trained on a synthetic dataset $\mathcal{D}_{s}$ for $N$ steps to generate a student trajectory ${\{\hat{\bm{\theta}}_t\}}_{t}^{t+N}$. This training process minimizes the cross-entropy loss $\mathcal{L}_{\text{ce}}$:
\begin{equation}
\hat{\bm{\theta}}_{t+i+1}=\hat{\bm{\theta}}_{t+i}-\alpha \nabla \mathcal{L}_{\text{ce}}(\mathcal{D}_s;\hat{\bm{\theta}}_{t+i}) \,,
\end{equation}
where $\alpha$ is the learning rate.
\par
The synthetic dataset $\mathcal{D}_s$ is optimized by minimizing the following matching loss:
\begin{equation}
\label{eq:tm}
\mathcal{L}_{\text{tm}} = \frac{{\Vert \hat{\bm{\theta}}_{t+N}-\bm{\theta}_{t+M} \Vert_2^2}}{{\Vert \bm{\theta}_{t+M}-\bm{\theta}_{t} \Vert_2^2}} \,.
\end{equation}
\par

\begin{figure}[t]
	\centering
	\includegraphics[width=\linewidth]{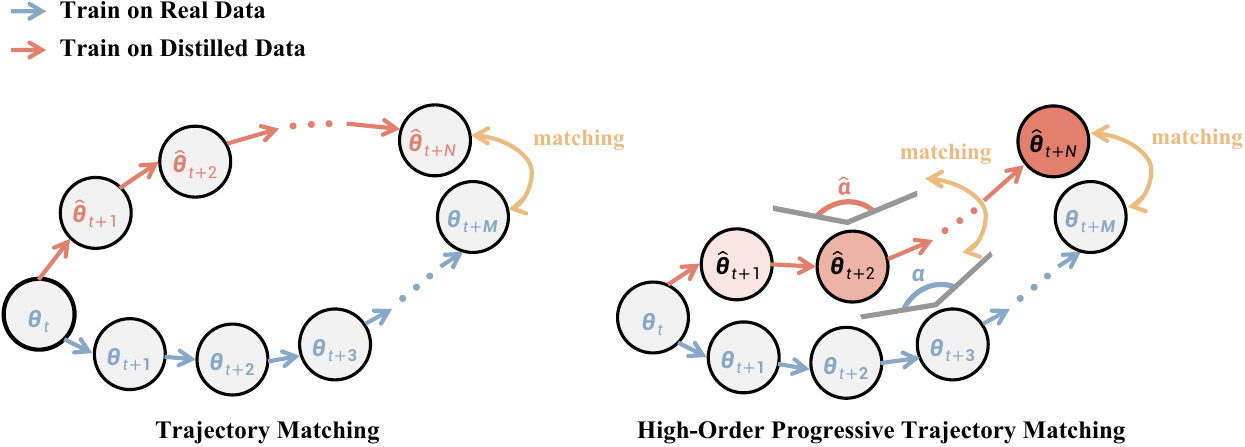}
	\vspace{-3mm}
	\caption{Comparison of trajectory matching approaches. Left: Conventional trajectory matching method. Right: Our proposed high-order progressive trajectory matching approach.}
	\label{overview}
	%\vspace{-6mm}
\end{figure}

\subsection{High-Order Trajectory Matching}
Current trajectory matching approaches for dataset distillation focus solely on comparing terminal states between expert and student trajectories, potentially missing valuable information about how parameters evolve during training. We propose matching trajectories with a high-order potential, which captures the geometric structure of parameter trajectories to synthesize datasets that better emulate the learning dynamics of training on the original data.
\par
Given sampled start parameters $\bm{\theta}_t$ and target parameters $\bm{\theta}_{t+M}$, we first obtain intermediate state parameters at training epoch $t+\lfloor \frac{M}{2} \rfloor$, denoted as $\bm{\theta}_{t+\lfloor \frac{M}{2} \rfloor}$, where $\lfloor \cdot \rfloor$ represents the mathematical floor function. Similarly, we compute intermediate student parameters $\hat{\bm{\theta}}_{t+\lfloor \frac{N}{2} \rfloor}$. To capture trajectory geometry, we introduce a high-order potential function that measures the angle formed by parameters at three time steps:
\begin{equation}
\begin{aligned}
&\psi(\bm{\theta}_i, \bm{\theta}_j, \bm{\theta}_k) = \cos \angle \bm{\theta}_i \bm{\theta}_j \bm{\theta}_k = \langle \bm{e}^{ij}, \bm{e}^{kj} \rangle \,, \\
&\text{where } \bm{e}^{ij}=\frac{\bm{\theta}_i-\bm{\theta}_j}{\Vert \bm{\theta}_i-\bm{\theta}_j \Vert_2} \,, \bm{e}^{kj}=\frac{\bm{\theta}_k-\bm{\theta}_j}{\Vert \bm{\theta}_k-\bm{\theta}_j \Vert_2} \,.
\end{aligned}
\end{equation}
\par
We then match geometric structures of real-data and synthetic-data trajectories using:
\begin{equation}
\mathcal{L}_{\text{hotm}} = \mathcal{L}_{1}(\psi(\bm{\theta}_t, \bm{\theta}_{t+\lfloor \frac{M}{2} \rfloor}, \bm{\theta}_{t+M}), \psi(\bm{\theta}_t, \hat{\bm{\theta}}_{t+\lfloor \frac{N}{2} \rfloor}, \hat{\bm{\theta}}_{t+N})) \,,
\end{equation}
where $\mathcal{L}_1$ is a smooth $\ell_1$ loss that is less sensitive to outliers~\cite{girshick2015fast}. 

\subsection{Easy-to-Complex Matching}
While high-order trajectory matching improves geometric alignment, we observe that parameter matching difficulty varies significantly across different network components. Some student parameters align easily with their teacher counterparts, while others require more complex optimization. To address this challenge, we propose an adaptive optimization strategy that progressively matches parameters from easy to complex, thereby improving the quality of the synthesized dataset.
\par
Formally, we define per-parameter losses $\ell_i$ derived from $\mathcal{L}_{\text{tm}}$ (see Eq. (\ref{eq:tm})) and introduce binary variables $\bm{v}=[v_1, v_2, \ldots, v_K]$, where $K$ represents the number of parameters to be matched. $\ell_i$ preserves the functional form of $\mathcal{L}_{\text{tm}}$ but applies it at the individual parameter level rather than across the entire parameter set. The synthetic dataset $\mathcal{D}_{s}$ and binary variables are jointly optimized by minimizing:
\begin{equation}
\underset{\mathcal{D}_s, \bm{v}}{\min}\mathbb{E}(\mathcal{D}_s,\bm{v},\kappa)=\sum_{i=1}^{K}v_i\ell_i-\kappa\sum_{i=1}^{K}v_i,\text{ s.t. } v_i\in [0,1] \,,
\label{eq}
\end{equation}
where $\kappa$ is a threshold parameter for controlling learning pace. Eq. (\ref{eq}) indicates that each parameter matching loss $\ell_i$ is discounted by a weight $v_i$. We use alternative convex search to solve Eq. (\ref{eq}). Specifically, at each distillation iteration, $\mathcal{D}_s$ and $\bm{v}$ are optimized alternately. When $\mathcal{D}_s$ is fixed, the optimal $\bm{v}^*$ can be determined by:
\begin{equation}
v_i^* =
\begin{cases}
1, & \text{if } \ell_i < \kappa \,, \\
0, & \text{otherwise} \,.
\end{cases}
\label{easy}
\end{equation}
This means that parameters yielding loss values below $\kappa$ are classified as easy parameters, while those exceeding $\kappa$ exhibit high matching complexity.
\par
During the optimization of $\mathcal{D}_s$ with fixed $\bm{v}$, only easy parameters are selected, and their corresponding loss values contribute to the update of $\mathcal{D}_s$.
\par
As the distillation progresses, we incrementally increase $\kappa$ every $P$ iterations by a growth factor $\mu$ ($\mu>1$):
\begin{equation}
\label{th}
\kappa=\kappa_{\text{base}}\mu^{\lfloor \frac{i}{P} \rfloor} \,.
\end{equation}
\par
This gradual increase in $\kappa$ enables the incorporation of increasingly complex parameters into the optimization of the synthetic dataset.

\subsection{Overall Loss}
Combining our high-order trajectory matching with the easy-to-complex strategy, we optimize the synthetic dataset using the following loss:
\begin{equation}
\mathcal{L}=\sum_{i=1}^{K}v_i\ell_i-\kappa\sum_{i=1}^{K}v_i+\lambda\mathcal{L}_{\text{hotm}} \,,
\label{loss}
\end{equation}
where $\lambda$ is a balancing hyper-parameter. This unified loss function enables our method to simultaneously leverage geometric trajectory information while adaptively focusing on manageable parameters during optimization. The pseudocode of our method is shown in Algorithm~\ref{alg}.

\begin{algorithm}[t]
     \caption{Pseudocode of Our Method}
     \label{alg}
     \KwIn{
        $\{\bm{\tau}\}$: Set of expert parameter trajectories \\
        $M$: Number of updates between initial and target expert parameters \\
        $N$: Number of student network updates per distillation iteration \\
        $T^{-}$, $T^{+}$: Sampling epoch range \\
        $\mu$: Growing factor \\
        $P$: Number of iterations for threshold $\kappa$ increment.
    }
     Initialize synthetic dataset $\mathcal{D}_{s}$ with random samples \\
     Initialize threshold $\kappa=\kappa_{\text{base}}$\\
     Initialize latent weights $\bm{v}=\bm{0}$ \\
     \For{each distillation iteration}{
     Sample trajectory $\bm{\tau}\in\{\bm{\tau}\}$ with $\bm{\tau}={\{\bm{\theta}_t\}}_{0}^{T}$ \\
     Sample starting epoch where $T^{-}\leq t \leq T^{+}$ \\
     Initialize student network $\hat{\bm{\theta}}_t = \bm{\theta}_t$ \\
     \For{$i=0$ to $N-1$}{
     Update student network: \\
    $\hat{\bm{\theta}}_{t+i+1}=\hat{\bm{\theta}}_{t+i}-\alpha \nabla \mathcal{L}_{\text{ce}}(\mathcal{D}_s;\hat{\bm{\theta}}_{t+i})$
     }
     Update $\kappa$ using Eq. (\ref{th}) and $\bm{v}$ using Eq. (\ref{easy})\\
     Obtain intermediate parameters $\bm{\theta}_{t+\lfloor \frac{M}{2} \rfloor}$ and $\hat{\bm{\theta}}_{t+\lfloor \frac{N}{2} \rfloor}$\\
     Compute loss with high-order matching using Eq. (\ref{loss})\\
     Update $\mathcal{D}_{s}$ and learning rate $\alpha$
     }
     
     \KwOut{Distilled dataset $\mathcal{D}_s$} 

\end{algorithm}

\begin{table}[t]
	\centering
	\addtolength{\tabcolsep}{2pt}
	\caption{Performance comparison of different methods on the PathMNIST dataset.}
  \vspace{-2mm}
  \resizebox{0.9\textwidth}{!}{
	\begin{tabular}{l|ccccc}
		\hline
            \multirow{2}{*}{ } &\multicolumn{5}{c}{\raisebox{-1pt}{IPC}}\\
            \cline{2-6}
        %& \multicolumn{5}{c}{PathMNIST} \\
		& \raisebox{-1pt}{1} & \raisebox{-1pt}{5} & \raisebox{-1pt}{10} & \raisebox{-1pt}{100} & \raisebox{-1pt}{1000} \\
		% \cline{2-6}
            \hline
		\raisebox{-1pt}{Real Dataset} & \multicolumn{5}{c}{\raisebox{-1pt}{89.89±0.49}} \\
        \hline
        DM~\cite{zhao2023dataset} & 38.39±4.39 & 62.85±0.73 & 66.99±1.04 & 82.04±0.88 & 87.29±0.59 \\
        IDM~\cite{zhao2023improved} &\textbf{50.39±0.53} & 69.32±1.62 & 72.74±1.13 & 82.05±0.89 & 87.51±0.21\\
        MTT~\cite{cazenavette2022dataset} &29.84±1.06 & 47.30±0.37 & 60.74±1.03 & 82.90±0.47 & 87.73±0.27\\
        FTD~\cite{du2023minimizing} & 29.36±0.77 & 55.99±1.02 & 62.06±0.93 & 82.81±0.88 & 87.65±0.39\\
		DATM~\cite{guo2023towards} & 45.74±1.66 & 64.94±1.01 & 73.18±0.90 & 84.07±1.03 & 89.15±0.11\\
        ATT~\cite{liu2025dataset} & 48.42±2.29 & 56.95±1.75 & 68.92±1.09 &83.86±0.67 & 88.41±0.32\\
        Ours & 47.71±1.22 & \textbf{73.92±0.66} & \textbf{77.23±0.65} & \textbf{84.82±0.40} & \textbf{89.86±0.36}\\
		\hline
	\end{tabular}%
        }
	\label{results1}%
 \vspace{-1mm}
\end{table}%

\begin{table*}[h]
	\centering
	\addtolength{\tabcolsep}{3pt}
	\caption{Performance comparison of different methods on the COVID19-CXR dataset.}
  \vspace{-2mm}
  \resizebox{0.8\textwidth}{!}{
	\begin{tabular}{l|cccc}
		\hline
            \multirow{2}{*}{ } &\multicolumn{4}{c}{\raisebox{-1pt}{IPC}}\\
        %&  \multicolumn{4}{c}{COVID19-CXR} \\
            \cline{2-5}
		& \raisebox{-1pt}{1} & \raisebox{-1pt}{5} & \raisebox{-1pt}{10} & \raisebox{-1pt}{50}\\
		% \cline{2-5}
            \hline
		\raisebox{-1pt}{Real Dataset}  & \multicolumn{4}{c}{\raisebox{-1pt}{91.49±0.42}}   \\
        \hline
        DM~\cite{zhao2023dataset} & 50.85±1.86 & 65.13±0.92 & 72.68±3.54 & 79.10±1.52\\
        IDM~\cite{zhao2023improved} & 67.78±0.92 & 68.51±0.92 & 72.77±0.82 & 79.28±0.94\\
        MTT~\cite{cazenavette2022dataset} & 56.98±1.89 & 74.94±1.74 & 83.30±0.17 & 86.11±0.52\\
        FTD~\cite{du2023minimizing} & 60.55±2.08 & 74.14±0.90 & 84.09±0.46 & 86.96±0.44\\
        DATM~\cite{guo2023towards} & 68.46±0.95 & 79.59±0.34 & 82.11±0.53 & 87.38±0.31\\
        ATT~\cite{liu2025dataset} & 67.83±2.02 & 73.80±0.92 &75.12±0.72 &87.62±0.44 \\
        Ours & \textbf{69.71±0.87} & \textbf{80.81±0.38} &\textbf{84.42±0.26} & \textbf{87.71±0.24} \\
		
	\hline
	\end{tabular}%
    }
	\label{results2}%
 \vspace{-1mm}
\end{table*}%

\section{Experiments}
\subsection{Datasets} 
We evaluate our method on two public medical datasets: PathMNIST~\cite{yang2023medmnist} and COVID19-CXR~\cite{rahman2021exploring}. PathMNIST contains 107,180 histopathology images of colon biopsies across 9 categories, split into 89,996/10,004/7,180 images for training, validation, and testing, respectively. COVID19-CXR comprises chest X-ray images with 3,616 COVID-19 cases, 6,012 lung opacity cases, 10,192 normal cases, and 1,345 viral pneumonia cases. We use an 8:2 train-test split for this dataset.

\subsection{Evaluation Metrics}
We evaluate distillation performance using classification accuracy. For each IPC (images per class) setting, we synthesize distilled datasets and train five networks with different random initializations, reporting the mean classification accuracy and standard deviation.

\subsection{Implementation Details}
Following prior work~\cite{zhao2023dataset,cazenavette2022dataset,du2023minimizing,guo2023towards,liu2025dataset}, we employ 3-layer and 5-layer ConvNets~\cite{gidaris2018dynamic} as our distillation networks for PathMNIST and COVID19-CXR, respectively. We train 100 teacher networks to obtain pre-recorded expert trajectories. Moreover, we set $P=500$.
\par
For PathMNIST, we resize images to 32$\times$32 pixels and set $\kappa_{\text{base}}=0.5$ and $\mu=1.5$. We use $\kappa_{\text{base}}=5\times 10^{-9}$ for $\text{IPC}=1,5,10$ and $\kappa_{\text{base}}=5\times 10^{-8}$ for $\text{IPC}=100,1000$. For COVID19-CXR, images are resized to 112$\times$112 with $\lambda=1.0$ and $\kappa_{\text{base}}=1\times 10^{-10}$. We set $\mu=1.3$ for $\text{IPC}=1,5,10$ and $\mu=1.6$ for $\text{IPC}=50$. We follow~\cite{guo2023towards} for the sampling parameters $t$, $M$, and $N$. All experiments are conducted on NVIDIA RTX 4090 GPUs.

\subsection{Comparison with State-of-the-Art}

As shown in Table~\ref{results1}, our method achieves superior performance across all settings on PathMNIST except $\text{IPC}=1$. Notably, with $\text{IPC}=1000$ (approximately 10\% of the original dataset size), our method approaches the accuracy achieved when training on the full dataset. On COVID19-CXR (Table~\ref{results2}), we consistently outperform all baselines across different IPC settings. The improvements over other trajectory matching-based methods demonstrate the effectiveness of our high-order potential and easy-to-complex matching strategy.

\begin{figure}[t]
	\centering
        \subfigure[PathMNIST]{
            \includegraphics[width=1\linewidth]{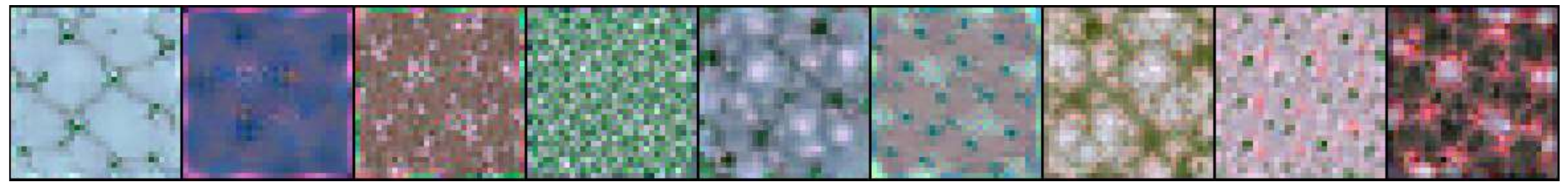}
        }
        \subfigure[COVID19-CXR]{
            \includegraphics[width=0.45\linewidth]{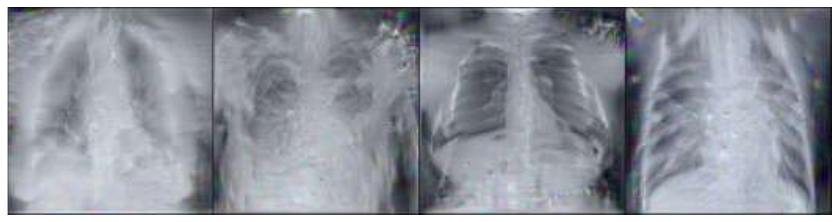}
        }
	\caption{Visualization of synthetic images generated through our method on two datasets with $\text{IPC}=1$.}
	\label{distilled_images}
\end{figure}

Fig.~\ref{distilled_images} visualizes distilled images from both datasets. The synthesized images exhibit highly abstract characteristics while maintaining discriminative features, effectively achieving data anonymization through information reduction.

\subsection{Discussion}
In this section, we investigate the generalization capability of our distilled images, analyze the effects of the proposed strategies, and examine the impact of hyperparameter choices.

\subsubsection{Cross-Architecture Generalization}
\begin{table*}[t]
	\centering
	\addtolength{\tabcolsep}{2pt}
	\caption{Cross-architecture generalization results on the PathMNIST dataset with $\text{IPC}=10$.}
  \vspace{-2mm}
        \resizebox{0.75\textwidth}{!}{
	\begin{tabular}{l|cccc}
		\hline
        \multicolumn{1}{l|}{\multirow{2}{*}{\raisebox{-1pt}{Method}}} & \multicolumn{4}{c}{\raisebox{-1pt}{Evaluation Model}} \\
        \cline{2-5}
		 & \raisebox{-1pt}{ConvNet~\cite{gidaris2018dynamic}} & \raisebox{-1pt}{ResNet18~\cite{he2016deep}} & \raisebox{-1pt}{VGG11~\cite{simonyan2014very}} & \raisebox{-1pt}{AlexNet~\cite{krizhevsky2012imagenet}}\\
        \hline
        DM~\cite{zhao2023dataset} & 66.99±1.04& 61.66±1.75 & 54.68±1.38 & 44.30±4.37\\
        IDM~\cite{zhao2023improved} & 72.74±1.13& 67.21±0.75 & 57.72±0.59 & 37.26±2.75\\
        MTT~\cite{cazenavette2022dataset} &60.74±1.03 & 58.67±0.70 & 46.87±3.04 & 42.44±4.66\\
        FTD~\cite{du2023minimizing} & 62.06±0.93 & 58.65±1.46 & 47.33±3.01 & 46.13±5.10 \\
		DATM~\cite{guo2023towards} & 73.18±0.90 & 60.36±1.78 & 49.61±3.21 & 47.67±1.84 \\
        ATT~\cite{liu2025dataset} & 68.92±1.09& 59.52±1.39 & 48.97±1.37 &34.14±1.77\\
        
        Ours & \textbf{77.23±0.65} & \textbf{68.36±1.52} & \textbf{59.61±2.06} & \textbf{48.85±4.44}\\
	\hline
	\end{tabular}%
    }
	\label{cross}%
 \vspace{-1mm}
\end{table*}%

We evaluate the generalization capability of our distilled images ($\text{IPC}=10$) on PathMNIST across three different architectures. Results in Table~\ref{cross} show that despite some performance degradation across architectures, our method consistently outperforms competing approaches, demonstrating strong generalization ability.

\begin{table*}[t]
	\centering
	\addtolength{\tabcolsep}{2pt}
	\caption{Ablation study on the proposed components using the PathMNIST dataset. HO: high-order trajectory matching; E2C: easy-to-complex matching.}
  \vspace{-2mm}
        \resizebox{0.9\textwidth}{!}{
	\begin{tabular}{>{\centering}m{0.8cm}>{\centering}m{0.8cm}|ccccc}
		\hline
          \multirow{2}{*}{\raisebox{-1pt}{HO}}  & \multirow{2}{*}{\raisebox{-1pt}{E2C}} &\multicolumn{5}{c}{\raisebox{-1pt}{IPC}}\\
            \cline{3-7}
        % & & \multicolumn{5}{c}{PathMNIST} \\
		 &  & \raisebox{-1pt}{1} & \raisebox{-1pt}{5} & \raisebox{-1pt}{10} & \raisebox{-1pt}{100} & \raisebox{-1pt}{1000} \\
        \hline
        & &29.84±1.06 & 47.30±0.37 & 60.74±1.03 & 82.90±0.47 & 87.73±0.27\\
        \checkmark&  & 47.03±2.61 & 73.14±0.39 & 76.77±0.62 & 84.28±0.53 & 89.58±0.43\\
		& \checkmark& 42.48±4.21 & 49.42±1.64 & 65.66±1.30 & 84.19±0.19 & 89.19±0.26\\
        
        \checkmark& \checkmark & \textbf{47.71±1.22} &\textbf{73.92±0.66} & \textbf{77.23±0.65} & \textbf{84.82±0.40} & \textbf{89.86±0.36}\\
		\hline
	\end{tabular}%
    }
	\label{component}%
 \vspace{-1mm}
\end{table*}%

\begin{table*}[t]
	\centering
	\addtolength{\tabcolsep}{2pt}
	\caption{Effect of hyperparameter $\lambda$ on the PathMNIST dataset.}
  \vspace{-2mm}
  \resizebox{0.85\textwidth}{!}{
	\begin{tabular}{>{\centering}m{1cm}|ccccc}
		\hline
         \multirow{2}{*}{\raisebox{-1pt}{$\lambda$}} &\multicolumn{5}{c}{\raisebox{-1pt}{IPC}}\\
         \cline{2-6}
        % & \multicolumn{5}{c}{PathMNIST} \\
	   & \raisebox{-1pt}{1} & \raisebox{-1pt}{5} & \raisebox{-1pt}{10} & \raisebox{-1pt}{100} & \raisebox{-1pt}{1000} \\
        \hline
        0 &29.84±1.06 & 47.30±0.37 & 60.74±1.03 & 82.90±0.47 & 87.73±0.27\\
         0.1 & 45.12±2.22 & 72.76±0.82 & \textbf{77.04±0.42} & 83.91±1.21 & 89.14±0.43\\
        0.5&  \textbf{47.03±2.61} &\textbf{73.14±0.39} & 76.77±0.62 & \textbf{84.28±0.53} & \textbf{89.58±0.43}\\
        1 & 46.59±3.36 & 72.92±0.85 & 76.67±0.71 & 84.10±0.91 & 89.32±0.37\\
        10 & 43.61±4.46 & 70.39±0.89 & 75.12±0.22 & 84.18±1.07 & 88.86±0.51\\
		\hline
	\end{tabular}%
	\label{weight}%
    }
 \vspace{-1mm}
\end{table*}%

 \subsubsection{Analysis of Matching Strategies}
Table~\ref{component} presents ablation studies on our two key matching strategies. The high-order trajectory matching yields significant improvements over the baseline by capturing richer geometric patterns in parameter trajectories. The easy-to-complex matching strategy further enhances performance across all IPC settings. When both strategies are combined, we achieve the best results.

\subsubsection{Hyper-Parameter Study}
We examine the impact of the high-order loss weight $\lambda$, excluding the easy-to-complex matching strategy to isolate its effect. As shown in Table~\ref{weight}, while different non-zero weights all improve performance compared to using only the original matching loss (Eq. (\ref{eq:tm})), an excessive weight (e.g., $\lambda=10$) can dominate the distillation process and limit gains. Based on empirical results, we set $\lambda=0.5$ for experiments.

\section{Conclusion}
In this paper, we introduce a high-order progressive trajectory matching approach for medical image dataset distillation. Our method's shape-wise potential captures geometric relationships in parameter trajectories, while the easy-to-complex matching strategy adaptively addresses parameters based on their optimization complexity. Experiments on various medical image classification tasks demonstrate that our approach achieves state-of-the-art performance in dataset distillation while maintaining privacy guarantees and model accuracy comparable to training on full datasets. This advancement enables efficient sharing of medical imaging data while respecting privacy constraints and reducing computational requirements. Future work could explore extending this approach to more complex medical imaging tasks such as segmentation and detection, as well as incorporating domain-specific medical knowledge into the distillation process. In addition, theoretical analysis of privacy guarantees could enhance the practical applicability of our approach in clinical settings.

%\newpage

\begin{credits}
\subsubsection{\ackname} 
The paper was supported by the Open Fund of Intelligent Control Laboratory 2024-ZKSYS-KF02-03.

\subsubsection{\discintname}
The authors have no competing interests to declare that are relevant to the content of this paper.
\end{credits}

\bibliographystyle{splncs04}
\bibliography{Paper-3876}




% LNCS Homepage, \url{http://www.springer.com/lncs}, last accessed 2023/10/25
% \end{thebibliography}
\end{document}